%% file: neurips_2024.tex
\documentclass{article}

\PassOptionsToPackage{numbers, sort&compress}{natbib}

\usepackage[final]{neurips_2024}

\usepackage[utf8]{inputenc} 
\usepackage[T1]{fontenc}    
\usepackage{hyperref}       
\usepackage{url}            
\usepackage{booktabs}       
\usepackage{amsfonts}       
\usepackage{nicefrac}       
\usepackage{microtype}      
\usepackage{xcolor}         
\usepackage{subcaption}
\usepackage{wrapfig}
\renewcommand{\paragraph}[1]{\textbf{#1} }

\usepackage{comment}
\usepackage{pgfplots}
\pgfplotsset{compat=1.18}

\usepackage{graphicx}
\usepackage{tikz}
\usepackage{amsmath}

\title{NARAIM: Native Aspect Ratio Autoregressive Image Models}

\author{%
  Daniel Gallo Fernández\thanks{Equal contribution} \\
  University of Amsterdam \\
  \texttt{daniel.gallo.fernandez@student.uva.nl} \\
  \And
  Robert van der Klis$^*$ \\
  University of Amsterdam \\
  \texttt{robert.van.der.klis@student.uva.nl} \\
  \AND
  Răzvan-Andrei Matişan$^*$ \\
  University of Amsterdam \\
  \texttt{razvan.matisan@student.uva.nl} \\
  \And
  Janusz Partyka$^*$ \\
  University of Amsterdam \\
  \texttt{janusz.partyka@student.uva.nl} \\
  \And
  Efstratios Gavves \\
  University of Amsterdam \\
  \texttt{e.gavves@uva.nl} \\
  \And
  Samuele Papa \\
  University of Amsterdam \\
  \texttt{s.papa@uva.nl} \\
  \And
  Phillip Lippe \\
  University of Amsterdam \\
  \texttt{p.lippe@uva.nl} \\
}

\begin{document}

\maketitle

\begin{abstract}
\input{sections/0_abstract}
\end{abstract}

\input{sections/1_intro}
\input{sections/2_method}
\input{sections/3_results}
\input{sections/4_conclusion}
\bibliography{references}
\bibliographystyle{bibstyle}
\newpage
\appendix
\onecolumn
\input{sections/appendix}

\end{document}

%% file: sections/0_abstract.tex
While vision transformers are able to solve a wide variety of computer vision tasks, no pre-training method has yet demonstrated the same scaling laws as observed in language models. Autoregressive models show promising results, but are commonly trained on images that are cropped or transformed into square images, which distorts or destroys information present in the input. To overcome this limitation, we propose NARAIM, a vision model pre-trained with an autoregressive objective that uses images in their native aspect ratio. By maintaining the native aspect ratio, we preserve the original spatial context, thereby enhancing the model's ability to interpret visual information. In our experiments, we show that maintaining the aspect ratio improves performance on a downstream classification task. 

%% file: sections/1_intro.tex
\section{Introduction}
\label{sec:intro}

Recent research has shown that pre-training large transformer models on vast datasets produces highly capable models, which can be fine-tuned for various tasks \cite{vaswani_attention_2017, brown2020language}. In language modelling, this is achieved by training models with billions of parameters on next-token prediction using datasets with trillions of tokens, as this prediction task scales well with data, compute, and model size \cite{brown2020language, touvron_llama_2023, dubey2024llama, reid2024gemini, kaplan_scaling_2020, hoffmann2022training}. However, in computer vision, no pre-training task has shown similarly favourable scaling laws \cite{el-nouby_scalable_2024}. For example, reconstruction tasks, e.g., masked autoencoders, have been found lacking when transferred to downstream tasks \cite{he_masked_2021, balestriero_learning_2024}. Inspired by the success of autoregressive objectives in language, \citet{el-nouby_scalable_2024} introduce Autoregressive Image Models (AIM), a class of vision transformers (ViTs) \cite{dosovitskiy2020image} trained with next patch prediction. These models achieve promising results, demonstrating similar scaling behaviours to autoregressive language models \cite{el-nouby_scalable_2024}.

While ViTs can efficiently train on images with varying aspect ratios, the common practice remains to resize images to a square resolution for training, typically using random resized cropping, which can distort image information while providing regularization \cite{el-nouby_scalable_2024,bardes2024revisiting,dehghani2023scaling,szegedy2015going}. However, \citet{dehghani_patch_2023} suggest that maintaining the native aspect ratio improves classification performance. We further hypothesize that this distortion is even more critical for generative pre-training objectives, where resizing could disrupt patterns or structures the model needs to reproduce.

Thus, in this paper, we propose Native Aspect Ratio Autoregressive Image Models (NARAIM), an AIM model that utilizes native aspect ratio inputs \cite{deng_imagenet_2009, dehghani_patch_2023, el-nouby_scalable_2024}. 
By preserving the aspect ratio during pre-training, our approach improves downstream classification accuracy without increasing computational costs, as the total number of input tokens remains constant.
Additionally, since images with varying aspect ratios require positional embeddings that adapt to different image layouts, we highlight the advantages of fractional positional embeddings in the pre-training process \cite{dehghani_patch_2023}. 
Finally, we demonstrate that random cropping while preserving the native aspect ratio serves as an effective regularizer, outperforming the standard random resized crop method in the original AIM models, particularly on inputs with highly non-square aspect ratios.

%% file: sections/2_method.tex
\section{Autoregressive Image Modeling with Native Aspect Ratios}
\label{sec:method}

\begin{figure*}[t!]
    \centering
    \includegraphics[width=1.0\textwidth]{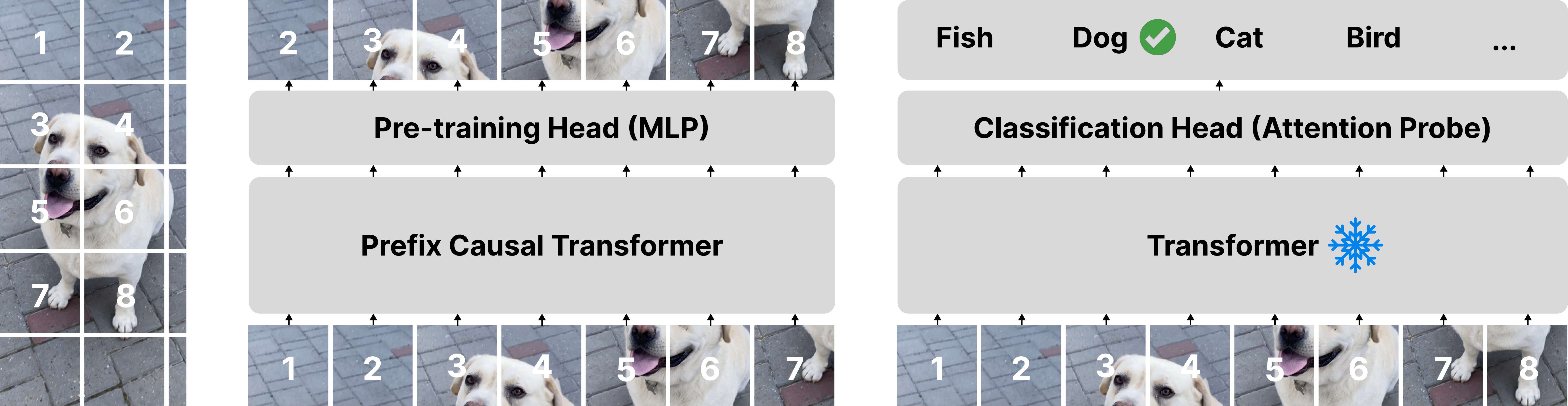}
    \caption{\textbf{NARAIM approach.} The input is divided into patches in row-major order, which are then processed by a vision transformer. The pre-training head utilizes the transformer's output to predict the next token based on the preceding ones. Meanwhile, the classification head, implemented with an attention probe, uses the transformer's output to predict a class.}
    \label{fig:model}
\end{figure*}

\begin{figure*}[t!]
    \centering
    \includegraphics[width=1.0\textwidth]{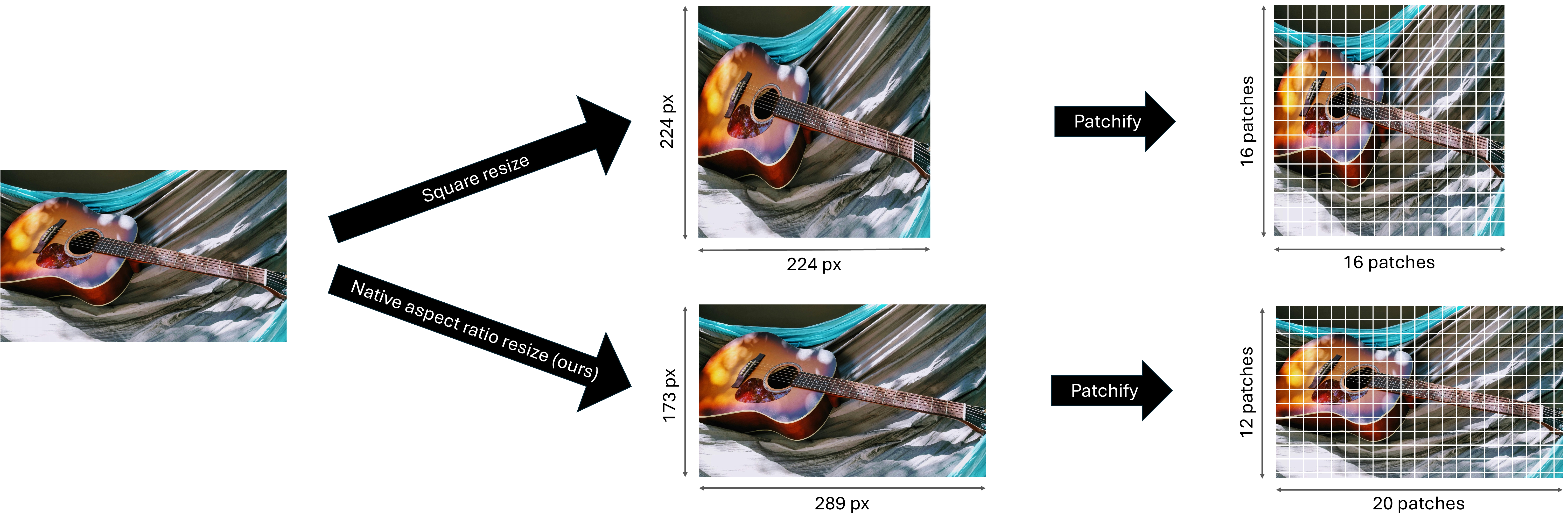}
    \caption{\textbf{Native aspect ratio resize.} Given a crop, it is common to resize it to a fixed-sized square. Since the image is going to be patchified and fed to a transformer, and the transformer itself is agnostic to the spatial organization of the patches, we propose keeping the native aspect ratio. First, we reshape the image keeping the aspect ratio fixed, ensuring the total number of pixels does not exceed $224^2$. Then, we patchify the image, obtaining at most 256 patches.}
    \label{fig:nar-sar-horizontally}
\end{figure*}

In this section, we introduce NARAIM: Native Aspect Ratio Autoregressive Image Models, a variant of Autoregressive Image Models (AIM) that maintains the native aspect ratio of input images.
AIM divides images into patches using a raster (row-major) ordering, which are subsequently processed by a ViT \citep{dosovitskiy2020image}.
Using a causal attention mask, the objective of the ViT is to predict the next image patch in pixel space, which results in a strong representation model which can be fine-tuned for downstream applications.
For a consistent shape, AIM resizes and crops images changing their aspect ratio, thus distorting potentially crucial image structures (see Figure \ref{fig:nar-sar-horizontally}). In NARAIM, we remove these distortions by using an aspect ratio preserving resize, as detailed below.

\paragraph{Native Aspect Ratio Resize.} Instead of forcing an image into a square shape of exactly $224^2$ pixels, we rescale the input image to an overall pixel number of approximately $224^2$, while keeping the aspect ratio unchanged. Next, we take a crop from the top-left such that the height and the width are some integer multiple of the patch size. After cropping, we split the image into square patches of size $14$ using a raster (row-major) ordering, which results in at most $256$ patches. The patches are then processed by the ViT and used to predict the next patch in pre-training, as shown in Figure~\ref{fig:model}, or other objectives like classification in downstream tasks later. Figure~\ref{fig:nar-sar-horizontally} illustrates the difference between the traditional pre-processing of inputs in square-sized images and the native aspect ratio resize. As images of different shapes can still produce small variations in the token number, we add padding tokens where needed, and mask them in the loss calculation.

\paragraph{Downstream Adaptation.} When the pre-training is completed, the model generates a feature vector for each patch. In order to evaluate the quality of these features, we freeze the ViT backbone and replace the pre-training head with a trainable attentive classification probe. We chose to use an attentive probe instead of a linear probe, as \citet{el-nouby_scalable_2024} found that the attentive probe performs substantially better. Moreover, we keep the input format to the models consistent, i.e., a model that was pre-trained on native aspect ratio inputs will also be used to classify native aspect ratio images. To more easily adapt the model to downstream tasks when causal attention is not needed anymore we randomly sample a prefix causal attention mask during pre-training, similar to \citet{el-nouby_scalable_2024}. We refer to Appendix~\ref{sec:prefix-causal-attention} for more details.

\paragraph{Augmentations.} The original AIM applied a random resized crop augmentation during training, which crops the image and changes its aspect ratio. While this augmentation helps to reduce overfitting, the distortions negatively impact the quality of the representations learned, as we show in Section~\ref{sec:experiments}. To maintain the regularization benefit, we adapt this augmentation to NARAIM by applying a random crop with the same aspect ratio as the original model, as explained in Figure~\ref{fig:nar-sar-horizontally}. In addition, we impose the constraint that the crop needs to contain at least $224^2$ pixels to prevent crops that would require severe upsampling. Additionally, we apply a random horizontal flip during training. During inference and downstream evaluation, we just take a native aspect ratio crop for NARAIM. For AIM, we follow \citet{el-nouby_scalable_2024} and first resize the image so that the shortest side is 256 pixels, and then we take a center crop with a side length of 224 pixels. 

\paragraph{Positional Embeddings.} Positional embeddings play a significant role in helping the model understand the spatial location of the patches. We experiment with two different ways of calculating the positional embeddings: absolute and fractional \cite{dehghani_patch_2023}. The key difference is that absolute positional embeddings use the \textit{index} of the patch along the horizontal and vertical axes, whereas fractional positional embeddings use the \textit{proportion}, that is, the index divided by the number of patches along that direction. For the absolute positional embedding, we take the height and width \textit{indices}, encode them using the fixed transformation from \citet{vaswani_attention_2017}, and concatenate both representations. For the fractional variant, we get the height and width \textit{proportions}, encode them using a learned dense layer, and sum them as in \citet{dehghani_patch_2023}. A mathematical formulation of how to compute these two positional embeddings can be found in Appendix~\ref{sec:positional-embeddings}.

We anticipate that absolute positional embeddings may cause problems when tested on images whose aspect ratio is out-of-distribution, whereas fractional positional embeddings are likely to generalize more easily. Moreover, fractional positional embeddings enable the model to recognize when a patch is near the edge (since the proportion will approach 1), whereas absolute positional embeddings do not provide this information, as the total number of patches is unknown.

\paragraph{Loss Functions.} We use two different loss functions for the pre-training head: with and without patch normalization. Let $\mathbf{x}_i \in \mathbb{R}^{3P^2}$ be the $i^{\text{th}}$ patch, where $P$ is the height and width of a patch. Then, with normalization, as proposed by \citet{he_masked_2021}, the $i^{\text{th}}$ prediction target $\mathbf{y}_i \in \mathbb{R}^{3P^2}$ is $\mathbf{y}_i = (\mathbf{x}_{i+1} -\mu(\mathbf{x}_{i+1})) / \sigma(\mathbf{x}_{i+1})$. Without normalization, the prediction target is simply $\mathbf{y}_i = \mathbf{x}_{i+1}$.


The loss function is the mean-squared error (MSE) between the predicted and ground-truth input: we sum up the squared differences per subpixel and divide by the total number of subpixels in the patch. For fine-tuning on the classification task, we use a standard cross-entropy loss function.

%% file: sections/3_results.tex
\section{Results and Discussion}
\label{sec:results}
\label{sec:experiments}

To demonstrate the benefits of native aspect ratios in NARAIM, we compare NARAIM to AIM on ImageNet-1k \cite{deng_imagenet_2009}, used for both the pre-training and the downstream tasks as in the small-scale studies of \citet{el-nouby_scalable_2024}. Our model is implemented in JAX/Flax \cite{jax2018github, flax2020github} and trained on a single 40GB Nvidia A100 GPU. Due to limited resources, our backbone is a ViT-B/14 with roughly 86M parameters and hence a scaled-down implementation of the original AIM model which had more than 600M parameters. 
Prior research \cite{dehghani_patch_2023} has shown that scaling models trained on native aspect ratio results in improved performance. We plan to verify that this scaling law extends also to our results in future work. Appendix \ref{sec:specifications} details the hyperparameters, and we make our code publicly available\footnote{\url{https://github.com/daniel-gallo/naraim}}.

\begin{table}[t!]
\caption{Validation MSE and accuracy of the pre-training and classification heads on ImageNet-1k with a ViT-B/14 backbone. The fine-tuning was repeated with four different seeds, and we report the mean and standard deviation. Using native aspect ratios in NARAIM improves the downstream classification accuracy over the original AIM \cite{el-nouby_scalable_2024} model. The results also highlight the importance of fractional embeddings, random crop augmentations, and normalization. In the first column, plus and minus signs are relative to the original NARAIM model (Appendix \ref{sec:specifications}).}
\centering
\label{tab:performance}
\begin{tabular}{lcc}
\toprule
\textbf{Model}             & \textbf{Next-Token MSE} & \textbf{Class Accuracy} \\
\midrule

AIM               & 0.340                       & $54.7 \pm 0.6$                 \\
NARAIM (ours)     & 0.357                       & $55.4 \pm 0.1$                 \\
\quad + Fractional embedding & 0.354                       & $56.0 \pm 0.1$                 \\
\quad + RandomCrop      & 0.354                       & $\textbf{56.8} \pm 0.1$                 \\
\quad - Normalization   & 0.010                       & $52.6 \pm 0.1$                \\ \bottomrule
\end{tabular}
\vspace{-5mm}
\end{table}

The results in Table \ref{tab:performance} demonstrate that NARAIM outperforms AIM in downstream performance. The best NARAIM model achieves a validation accuracy of $56.8 \pm 0.1$, surpassing AIM by more than two percentage points. While the pre-training validation MSE is lower for the AIM model, the ultimate goal of these models is to create useful representations for downstream tasks. To measure the usefulness of representations for downstream tasks, the downstream validation accuracy is a more useful proxy than the pre-training validation MSE. In Appendix~\ref{sec:appendix_raster}, we visualize the MSE across different patch locations in the image. Both methods show similar MSE patterns, with the borders being most difficult due to the patch discontinuity. 

\paragraph{Ablations.}
As for the ablation studies, we find that using fractional embeddings provides a small accuracy increase over absolute embeddings, likely due to the model with fractional embeddings being slightly better at classifying images whose aspect ratio is out-of-distribution. Using random crops substantially increases the downstream validation accuracy, which is explained by the fact that the base NARAIM model overfits during pre-training: thus, random crops alleviate this issue. Finally, looking at patch-wise normalization, we note that the pre-training MSE is on a different scale, and therefore cannot be compared. As for the downstream validation accuracy, we note that patch-wise normalization is very important. One intuitive explanation is this: if patches are not normalized, the pre-trained model will be geared towards detecting global patterns. Hence, the classifier may develop shortcuts such as predicting ``boat'' whenever it detects a blue patch, decreasing the accuracy.

\begin{wrapfigure}{r}{0.4\textwidth}
    \centering
    \vspace{-4mm}
    \includegraphics[width=\linewidth]{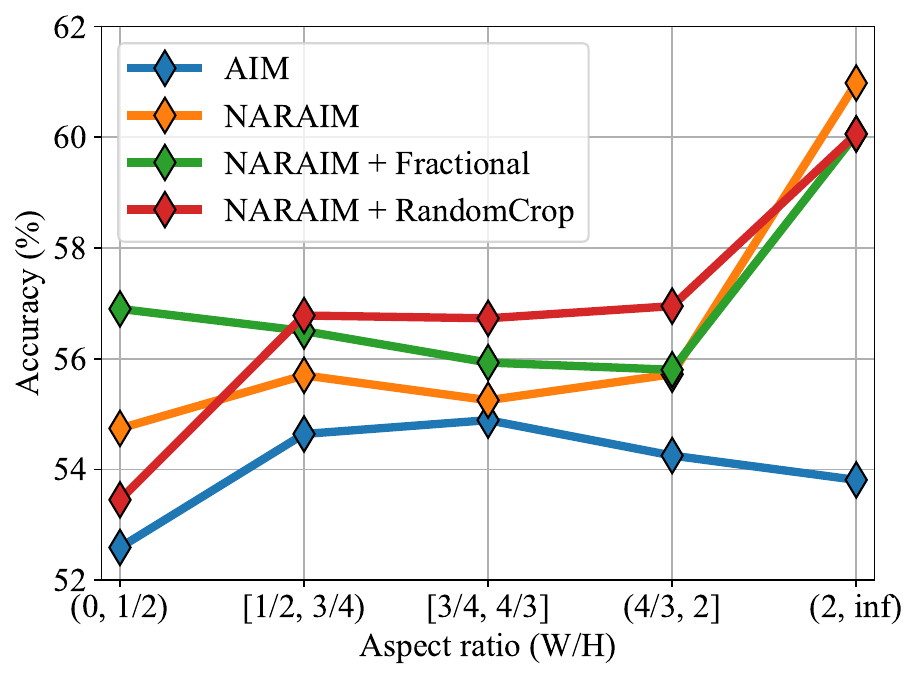}
    \caption{The classification accuracy over image aspect ratios. NARAIM improves across all aspect ratios.}
    \label{fig:aspect_ratio}
    \vspace{-4mm}
\end{wrapfigure}

\paragraph{Effect of Aspect Ratio.}
To gain further insights into the improved performance, we group images by their aspect ratios and visualize the classification accuracy per group in Figure~\ref{fig:aspect_ratio}. First, we note that results on the bins with aspect ratio $(0, 1/2)$ and $(2, \text{inf})$ are slightly noisy due to only containing several hundred images (see Appendix \ref{sec:aspect-ratio-distribution} for more information). Nonetheless, we observe that the accuracies for the NARAIM models are higher than that of the AIM model for every bin. We also note that the performance of the AIM model decreases when the model is applied to inputs with non-square aspect ratios, showing its focus on square-shaped images. The ``NARAIM" and ``NARAIM + Fractional" models showcase slightly noisy behavior over the bins, which may be attributed to the earlier observation that they overfit on the training set due to a lack of data augmentation.

In comparison, the performance for ``NARAIM + RandomCrop" model is remarkably stable in the middle three bins. Furthermore, we find a significant performance improvement for strongly horizontal images, though the precise gap may be noisy due to limited samples in this bin. Overall, we conclude that our ``NARAIM + RandomCrop" model is better at dealing with images with non-square aspect ratios than the ``AIM" model.

%% file: sections/4_conclusion.tex
\paragraph{Conclusion.} 
In this paper, we have introduced NARAIM, a vision transformer model pre-trained with an autoregressive objective that uses images in their native aspect ratio. We showed that using native aspect ratios improves the downstream classification performance, in particular for non-square images, emphasizing the importance of modeling images in their original aspect ratio in autoregressive image models. In future work, we plan to scale our model to 600 million parameters to benchmark it against the original AIM and validate its benefits at scale, as well as extending the downstream evaluation to more tasks.

%% file: sections/appendix.tex
\section{Aspect ratio distribution}
\label{sec:aspect-ratio-distribution}

\begin{figure}
    \centering
    \includegraphics[width=.75\linewidth]{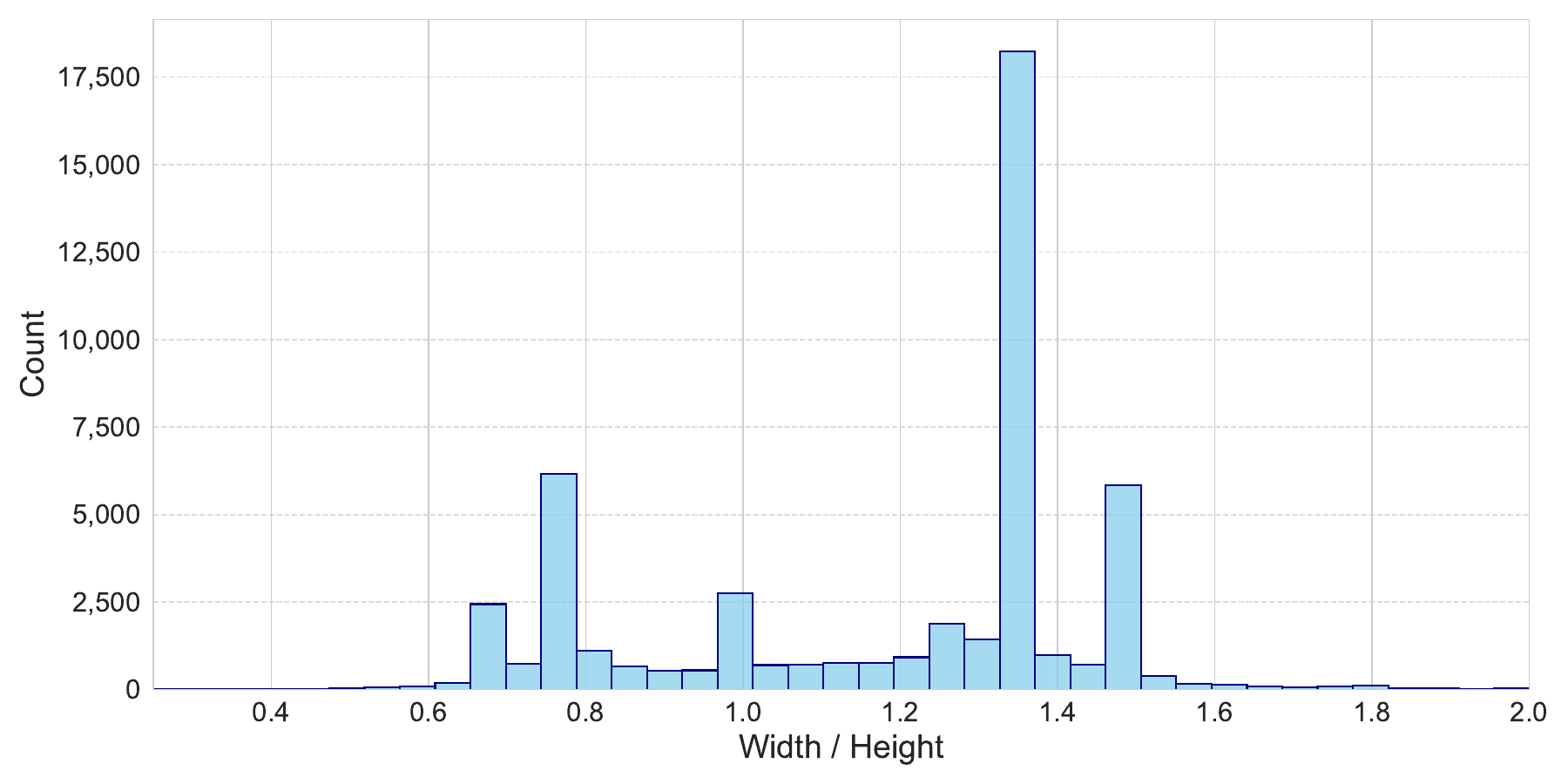}
    \caption{\textbf{Aspect ratio distribution.} This histogram shows the aspect ratios of the ImageNet-1k validation set. Most of the images are in landscape orientation, but we can observe three modes corresponding to portrait, square, and landscape images.}
    \label{fig:aspect-ratio-distribution}
\end{figure}

We use ImageNet-1k to train our models, both during pre-training and downstream classification. Figure \ref{fig:aspect-ratio-distribution} illustrates the aspect ratio of the images in the validation set. 

\section{NARAIM specifications}
\label{sec:specifications}
The baseline NARAIM model uses absolute positional embeddings, patch-wise normalization, and no random cropping. In Table \ref{tab:hp-backbone} we show the hyperparameters of the backbone and in Table \ref{tab:hp-training} we show the training parameters. 

\begin{table}[h]
    \caption{Hyperparameters of the backbone (ViT-B/14).}
    \label{tab:hp-backbone}
    
    \centering
    \begin{tabular}{ll}
    \textbf{Name}       & \textbf{Value} \\ \hline
    Patch size          & 14             \\
    Number of layers    & 12             \\
    Number of heads     & 12             \\
    Embedding dimension & 768            \\
    Hidden dimension    & 3,072          
    \end{tabular}
\end{table}

\begin{table}[h]
\caption{ Training parameters (very similar to \citet{el-nouby_scalable_2024}).}
\label{tab:hp-training}
\centering

\begin{tabular}{l|l|l}
\textbf{Parameter}                           & \textbf{Pre-training value}     & \textbf{Fine-tuning value} \\ \hline
Optimizer                                    & AdamW                           & AdamW                      \\     
Optimizer momentum                           & $\beta_1 = 0.9, \beta_2 = 0.98$ & $\beta_1 = 0.9$, $\beta_2 = 0.999$ \\   
Peak learning rate                           & $1e^{-3}$                       & $1e^{-3}$                       \\
Minimum learning rate                        & $0$                             & $1e^{-5}$                       \\ 
Weight decay                                 & $0.01$                          & $0.1$                        \\ 
Batch size                                   & $512$                           & $512$                        \\
Patch size                                   & $14 \times 14$                  & $14 \times 14$                     \\
Gradient clipping                            & $1.0$                           & $3.0$                             \\
Decay rate (for lr scheduler)                          & 0.1                   & -                        \\
Warmup iterations                            & 5,000                           & 500                   \\
Cooldown iterations                          & 10,000                          & -               \\    
Total iterations                             & 500,000                         & 50,000  \\
Learning rate schedule                       & exponential                     & cosine decay\\

\end{tabular}
\end{table}

\section{Positional embeddings}
\label{sec:positional-embeddings}

To calculate the fractional embedding for a patch $\mathbf{x}_i$, let $h_i, w_i$ be the vertical and horizontal indices of the patch $\mathbf{x}_i$ in the input, with $H, W$ the total number of vertical and horizontal patches in the input. Then, the fractional embedding for the patch is $f(h_i/H) + g(w_i/W)$, where $f$ and $g$ are learnable one-layer perceptrons.

Regarding the absolute embeddings, let  $h_i, w_i$ again be the vertical and horizontal indices, and let $\phi$ be the function introduced in the original Transformer paper \cite{vaswani_attention_2017}:
\[\phi(\text{pos}, 2i) = \sin(\text{pos}/10000^{2i/d}),\]
and
\[\phi(\text{pos}, 2i+1) = \cos(\text{pos}/10000^{2i/d}),\]
where $d = \frac{d_\text{model}}{2}$.
The complete positional embedding is then: 
\[\phi\left(h_i, 1 \colon d\right) \oplus \phi\left(w_i, 1 \colon d\right),\]
where $\oplus$ denotes concatenation.

\section{Prefix causal attention}
\label{sec:prefix-causal-attention}

Adapting a model trained with causal attention to downstream tasks may cause issues: due to the attention mask, the model only learns to create representations using causal attention. This means that for a token $\mathbf{x}_i$, only the information from tokens $\{\mathbf{x}_1, ..., \mathbf{x}_{i-1}\}$ will be incorporated. However, in downstream tasks, every token should be able to attend to every other token.

\begin{figure}[h]
  \begin{subfigure}{0.45\textwidth}\includegraphics[width=\columnwidth]{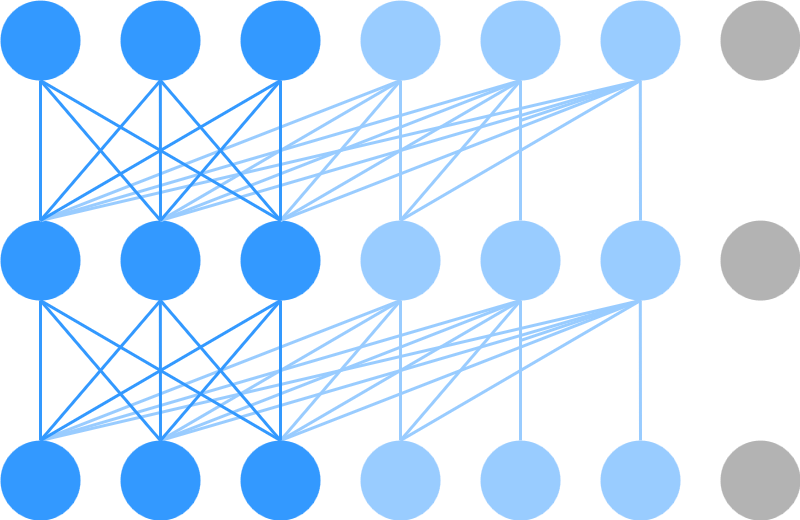}
  \end{subfigure}
  \hfill
  \begin{subfigure}{0.45\textwidth}\includegraphics[width=\columnwidth]{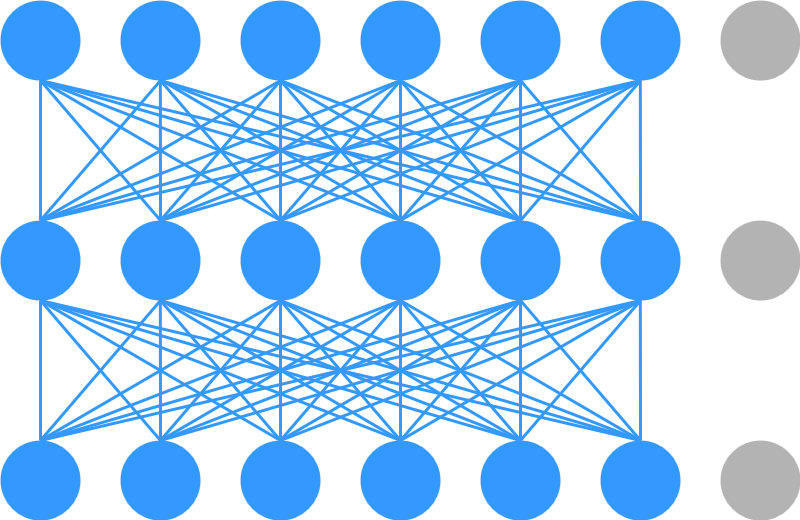}
  \end{subfigure}
  \caption{\textbf{Prefix causal attention.} For pre-training (left), we uniformly sample a prefix length $n$ during pre-training (e.g., $n = 3$). The attention for the first $n$ patches is set to be bidirectional and no loss will be computed for them. The rest of the patches adopt a causal mask and their loss is calculated. During fine-tuning to a downstream task (right), the mask is discarded. The gray patches represent the padding, which are added for reasons explained in Section~\ref{sec:method}.}
  \label{fig:prefix-causal-attention}
\end{figure}

Prefix causal attention addresses this issue by selecting a random integer $n$ between $1$ and $N-1$, where $N$ is the number of tokens in an input \cite{raffel_t5_2023}. Then, for tokens $\{\mathbf{x}_1, ..., \mathbf{x}_n\}$, the causal attention mask is dropped, and every token in this set can attend to every other token in the set. For the remaining tokens, $\{\mathbf{x}_{n+1}, ..., \mathbf{x}_N\}$, we use the standard causal attention mask. Predictions for tokens $\{\mathbf{x}_1, ..., \mathbf{x}_n\}$ are not included in the loss calculation, as the tokens could be trivially predicted. Figure \ref{fig:prefix-causal-attention} displays the prefix causal attention for pre-training and fine-tuning.

\citet{el-nouby_scalable_2024} trained networks with both prefix causal attention and regular causal attention, and found that prefix causal attention substantially improved the performance on downstream tasks.

\section{Raster patterns across patches}
\label{sec:appendix_raster}

\begin{figure}
    \begin{subfigure}{0.48\textwidth} \includegraphics[width=\columnwidth]{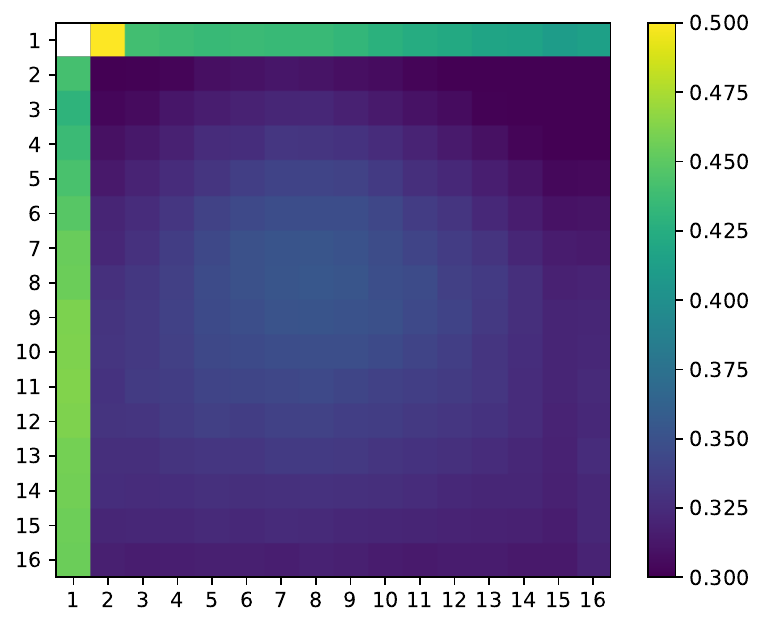}
    \subcaption{AIM}
    \label{fig:per-path-validation-mse-aim}
    \end{subfigure}
    \hfill
    \begin{subfigure}{0.48\textwidth} \includegraphics[width=\columnwidth]{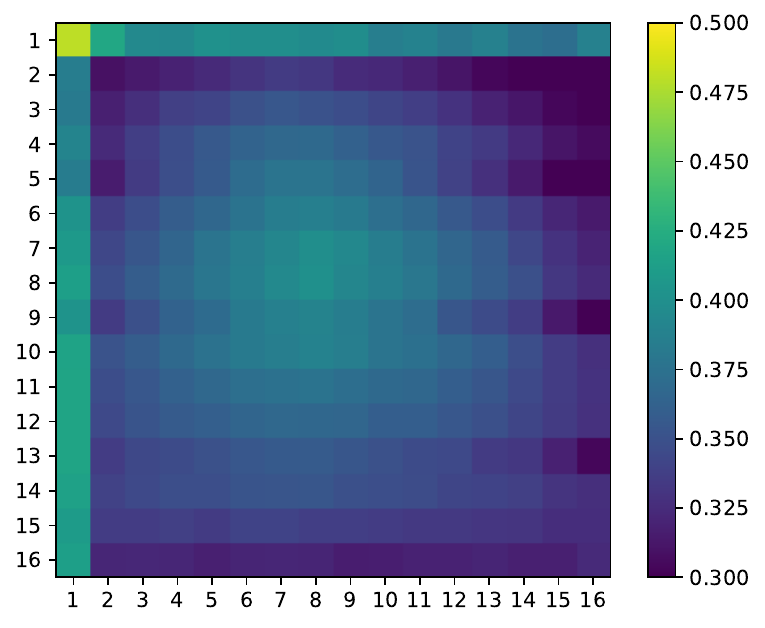}
    \subcaption{NARAIM with fractional embeddings}
    \label{fig:per-path-validation-mse-naraim}
    \end{subfigure}
    \caption{\textbf{Per-patch validation MSE.} In both models we see that predicting the first row and the first column is harder. Predicting the first row is hard because there is no information about the previous line. Predicting the first column is challenging because these patches are not correlated with the previous ones due to the carriage return.}
    \label{fig:per-patch-validation-mse}
\end{figure}

While \citet{el-nouby_scalable_2024} reports the validation MSE per-row, we report it per-patch in Figure \ref{fig:per-patch-validation-mse}. This is easy to compute for AIM, that always has 256 patches. The first one is never predicted, which is why it is white on Figure \ref{fig:per-path-validation-mse-aim}.

For NARAIM, we first map the patch index to the interval $[0, 1]$, and then we discretize this into 16 bins. The first patch is never predicted, but the second patch can lie in the first bin if the image was horizontal. Hence, the white square of Figure \ref{fig:per-path-validation-mse-aim} is not present in Figure \ref{fig:per-path-validation-mse-naraim}.

